\documentclass[10pt,twocolumn,letterpaper]{article}

\usepackage{cvpr}
\usepackage{times}
\usepackage{epsfig}
\usepackage{graphicx}
\usepackage{amsmath}
\usepackage{amssymb}
\usepackage{breqn}
\usepackage{algorithm,algorithmic}
\usepackage{paralist, tabularx}

\cvprfinalcopy 

\newcommand{\ignore}[1]{}

\def\cvprPaperID{****} 

\begin{document}

\title{Generative Face Completion}

\author{Yijun Li$^1$, ~Sifei Liu$^1$, ~Jimei Yang$^2$, ~and Ming-Hsuan Yang$^1$\\
$^1$University of California, Merced~~~~~~~~$^2$Adobe Research\\
{\tt\small \{yli62,sliu32,mhyang\}@ucmerced.edu ~~jimyang@adobe.com}
}

\maketitle

\begin{abstract}
In this paper, we propose an effective face completion algorithm using a deep generative model.
Different from well-studied background completion, the face completion task is more challenging as it often requires to generate semantically new pixels for the missing key components (e.g., eyes and mouths) that contain large appearance variations.
Unlike existing nonparametric algorithms that search for patches to synthesize, our algorithm directly generates contents for missing regions based on a neural network.
The model is trained with a combination of a reconstruction loss, two adversarial losses and a semantic parsing loss, which ensures
pixel faithfulness and local-global contents consistency.
With extensive experimental results,  we demonstrate
qualitatively and quantitatively that our model is able to deal with a large area of missing pixels in
arbitrary shapes and generate realistic face completion results.

\end{abstract}

\section{Introduction}

Image completion, as a common image editing operation, aims to fill the missing or masked regions in images with plausibly synthesized contents.
The generated contents can either be as accurate as the original, or simply fit well within the context such that the completed image appears to be visually realistic.
Most existing completion algorithms~\cite{barnes-2009-patchmatch, huang-2014-image} rely on low-level cues to search for patches from known regions of the same image and synthesize the contents that locally appear similarly to the matched patches.
These approaches are all fundamentally constrained to copy existing patterns and structures from the known regions.
The copy-and-paste strategy performs particularly well for background completion (e.g., grass, sky, and mountain) by removing foreground objects and filling the unknown regions with similar pattens from backgrounds.

\begin{figure}[t]
\centering
\footnotesize
{
\begin{tabular}{c@{\hspace{0.01\linewidth}}c@{\hspace{0.01\linewidth}}c@{\hspace{0.01\linewidth}}c@{\hspace{0.01\linewidth}}c@{\hspace{0.01\linewidth}}c@{\hspace{0.01\linewidth}}c@{\hspace{0.01\linewidth}}c@{\hspace{0.01\linewidth}}c@{\hspace{0.01\linewidth}}c}

\includegraphics[width = .264\linewidth]{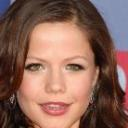} &
\includegraphics[width = .264\linewidth]{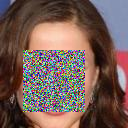} &
\includegraphics[width = .264\linewidth]{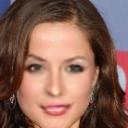} \\

\includegraphics[width = .264\linewidth]{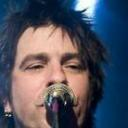} &
\includegraphics[width = .264\linewidth]{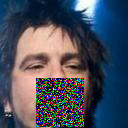} &
\includegraphics[width = .264\linewidth]{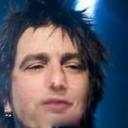} & \\

{(a) }& {(b) }& {(c)}\\
\end{tabular}
}
\caption{Face completion results. In each row from left to right: (a) original image (128 $\times$ 128 pixels). (b) masked input. (c) completion results by our method.
In the top row, the face is masked by a square. In the bottom row we show a real example where the mouth region is occluded by the microphone.}
\label{fig:problem}
\end{figure}

However, the assumption of similar patterns can be found in the same image
does not hold for filling missing parts of an object image (e.g., face).
Many object parts contain unique patterns, which cannot be matched
to other patches within the input image, as shown in Figure~\ref{fig:problem}(b).
An alternative is to use external databases as references~\cite{hays-2007-scene}.
Although similar patches or images may be found, the unique patterns of objects that involve semantic representation are not well modeled,
since both low-level~\cite{barnes-2009-patchmatch} and mid-level~\cite{huang-2014-image} visual cues of the known regions are not sufficient to infer semantically valid contents in missing regions.
In this paper, we propose an effective object completion algorithm using a deep generative model.
The input is first masked with noise pixels on randomly selected square region, and then fed into an autoencoder~\cite{vincent-2010-stacked}.
While the encoder maps the masked input to hidden representations, the decoder generates a filled image as its output.
We regularize the training process of the generative model by introducing two adversarial losses~\cite{goodfellow-2014-GAN}:
a local loss for the missing region to ensure the generated contents are semantically coherent, and a global one for the entire image to render more realistic and visually pleasing results.
%
In addition, we also propose a face parsing network~\cite{Liu_2015_CVPR,ParsingHelenLabel-2013-exemplar,ParsingHelen-2012-interactive} as an additional loss to regularize the generation procedure and enforce a more reasonable and consistent result with contexts.
%
%
This generative model allows fast feed-forward image completion without requiring an
external databases as reference.
For concreteness, we apply the proposed object completion algorithm on face images.

The main contributions of this work are summarized as follows.
First, we propose a deep generative completion model that consists of an encoding-decoding generator and two adversarial discriminators to synthesize the missing contents from random noise.
Second, we tackle the challenging face completion task and show the proposed model
is able to generate semantically valid patterns based on learned representations
of this object class.
Third, we demonstrate the effectiveness of semantic parsing in generation, which renders the completion results that look both more plausible and consistent with surrounding contexts.

\section{Related Work}

\paragraph{Image completion.} Image completion has been studied in
numerous contexts, e.g., inpainting, texture synthesis, and sparse signal recovery.
Since a thorough literature review is beyond the scope of this paper, and
we discuss the most representative methods to put our work in proper context.

An early inpainting method~\cite{bertalmio-2000-image} exploits
a diffusion equation to iteratively propagate
low-level features from known regions to unknown areas along the mask boundaries.
While it performs well on inpainting, it is limited to deal with small and homogeneous regions.
Another method has been developed to further improve inpainting results
by introducing texture synthesis \cite{bertalmio-2003-simultaneous}.
In~\cite{zora-2011-epll}, the patch prior is learned to restore images with missing pixels.
Recently Ren et al.~\cite{ren-2015-shepard} learn a convolutional network for inpainting.
The performance of image completion is significantly improved
by an efficient patch matching algorithm~\cite{barnes-2009-patchmatch}
for nonparametric texture synthesis.
While it performs well when similar patches can be found,
it is likely to fail when the source image does not contain
sufficient amount of data to fill in the unknown regions.
We note this typically occurs in object completion as
each part is likely to be unique and no plausible patches for the missing region can be found.
Although this problem can be alleviated by using an external database~\cite{hays-2007-scene},
the ensuing issue is the need to learn high-level representation
of one specific object class for patch match.

Wright et al.~\cite{wright-2009-sparseface} cast image completion as the task for recovering
sparse signals from inputs.
By solving a sparse linear system, an image can be recovered from some corrupted input.
However, this algorithm requires the images to be highly-structured
(i.e., data points are assumed to lie in a low-dimensional subspace),
e.g., well-aligned face images.
In contrast, our algorithm is able to perform object completion without strict constraints.

\paragraph{Image generation.}
Vincent et al.~\cite{vincent-2008-Denosing} introduce denoising autoencoders that learn to reconstruct clean signals from corrupted inputs.
In~\cite{dosovitskiy-2016-inverting}, Dosovitskiy et al. demonstrate that an object image can be reconstructed by inverting deep convolutional network features
(e.g., VGG~\cite{simonyan-2014-VGG}) through a decoder network.
Kingma et al.~\cite{kingma-2013-VAE} propose variational autoencoders (VAEs)
which regularize encoders by imposing prior over the latent units such that images can be generated by sampling from or interpolating latent units.
However, the generated images by a VAE are usually blurry due to its training objective based on
pixel-wise Gaussian likelihood.
Larsen et al.~\cite{larsen-2016-autoencoding} improve a VAE by adding a discriminator
for adversarial training which stems from the generative adversarial networks (GANs)~\cite{goodfellow-2014-GAN} and demonstrate
more realistic images can be generated.

\begin{figure*}[t]
\begin{center}
\includegraphics[width=17cm]{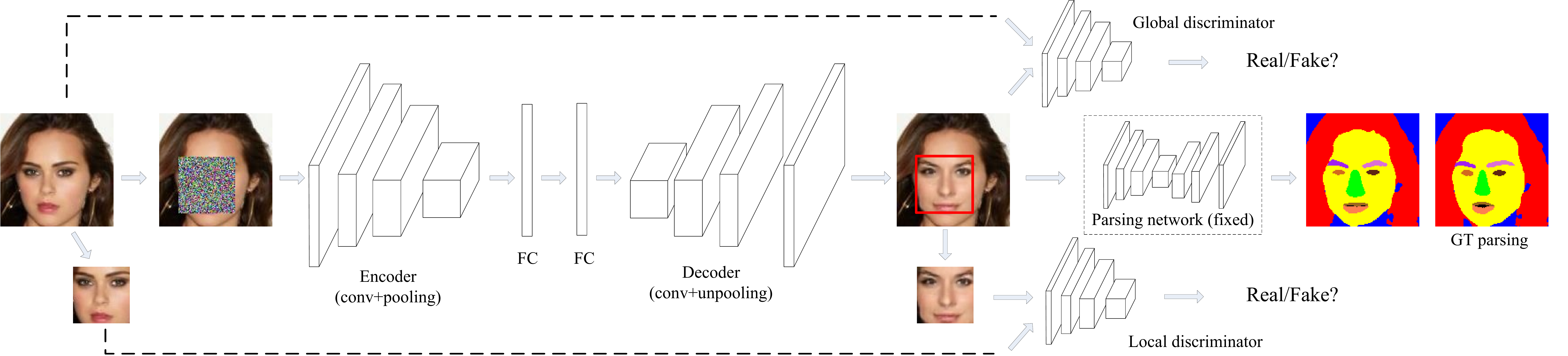}
\end{center}
\caption{Network architecture. It consists of one generator, two discriminators and a parsing network. The generator takes the masked image as input and outputs the generated image. We replace pixels in the non-mask region of the generated image with original pixels.
Two discriminators are learned to distinguish the synthesize contents in the mask and whole generated image as real and fake.
The parsing network, which is a pretrained model and remains fixed, is to further
ensure the new generated contents more photo-realistic and encourage
consistency between new and old pixels.
Note that only the generator is needed during the testing.}
\label{fig:framework}
\end{figure*}

Closest to this work is the method proposed by Deepak et al.~\cite{Deepak-2016-ContextEncoder} which applies an autoencoder and
integrates learning visual representations with image completion.
However, this approach emphasizes more on unsupervised
learning of representations than image completion.
In essence, this is a chicken-and-egg problem.
Despite the promising results on object detection, it is still not entirely clear
if image completion can provide sufficient supervision signals for learning high-level features.
On the other hand, semantic labels or segmentations
are likely to be useful for improving the completion results,
especially on a certain object category.
With the goal of achieving high-quality image completion, we propose to use
an additional semantic parsing network to regularize the generative networks.
Our model deals with severe image corruption (large region with missing pixels), and develops a combined reconstruction, adversarial and parsing loss for face completion.

\section{Proposed Algorithm}

In this section, we describe the proposed model for object completion.
Given a masked image, our goal is to synthesize the missing contents that are both semantically
consistent with the whole object and visually realistic.
Figure~\ref{fig:framework} shows the proposed network that consists of one generator, two discriminators, and a parsing network.

\subsection{Generator}

The generator $\mathcal G$ is designed as an autoencoder to construct
new contents given input images with missing regions.
The masked (or corrupted) input, along with the filled noise, is first mapped to hidden representations through the encoder.
Unlike the original GAN model~\cite{goodfellow-2014-GAN} which directly starts from a noise vector, the hidden representations obtained from the encoder capture more variations and relationships between unknown and known regions, which are then fed into the decoder
for generating contents.

We use the architecture from ``conv1" to ``pool3" of the VGG-19~\cite{simonyan-2014-VGG} network, stack two more convolution layers and one more pooling layer on top of that,
and add a fully-connected layer after that as the encoder.
The decoder is symmetric to the encoder with unpooling layers.

\subsection{Discriminator}
\begin{figure*}[t]
\centering
\small
{
\begin{tabular}{c@{\hspace{0.01\linewidth}}c@{\hspace{0.01\linewidth}}c@{\hspace{0.01\linewidth}}c@{\hspace{0.01\linewidth}}c@{\hspace{0.01\linewidth}}c@{\hspace{0.01\linewidth}}c@{\hspace{0.01\linewidth}}c}
\includegraphics[width = .132\linewidth]{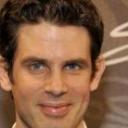} &
\includegraphics[width = .132\linewidth]{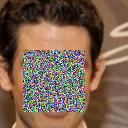} &
\includegraphics[width = .132\linewidth]{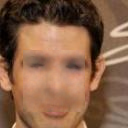} &
\includegraphics[width = .132\linewidth]{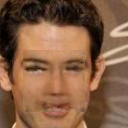} &
\includegraphics[width = .132\linewidth]{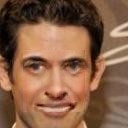} &
\includegraphics[width = .132\linewidth]{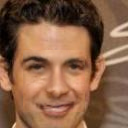} &
\includegraphics[width = .132\linewidth]{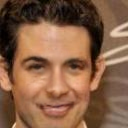} & \\

\includegraphics[width = .132\linewidth]{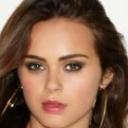} &
\includegraphics[width = .132\linewidth]{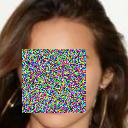} &
\includegraphics[width = .132\linewidth]{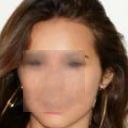} &
\includegraphics[width = .132\linewidth]{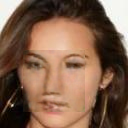} &
\includegraphics[width = .132\linewidth]{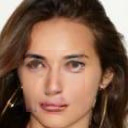} &
\includegraphics[width = .132\linewidth]{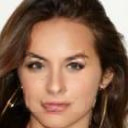} &
\includegraphics[width = .132\linewidth]{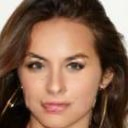} & \\

{(a) Original image}& {(b) Masked input}& {(c) M1}& {(d) M2}& {(e) M3}& {(f) M4}& {(g) M5}\\
\end{tabular}
}
\caption{Completion results under different settings of our model. (c) M1: $L_{r}$. (d) M2: $L_{r} + L_{a_1}$. (e) M3: $L_{r} + L_{a_1} + L_{a_2}$. (f) M4: $L_{r} + L_{a_1} + L_{a_2} + L_{p}$. The result in (f) shows the most realistic and plausible completed content. It can be further improved through post-processing techniques such as (g) M5: M4 $+$ Poisson blending~\cite{PB-2003-poisson} to eliminate subtle color difference along mask boundaries.}
\label{fig:step}
\end{figure*}

The generator can be trained to fill the masked region or missing pixels
with small reconstruction errors.
However, it does not ensure that the filled region is visually realistic and coherent.
As shown in Figure~\ref{fig:step}(c), the generated pixels are quite blurry and only capture the coarse shape of missing face components.
To encourage more photo-realistic results,
we adopt a discriminator $\mathcal D$ that serves as a binary classifier to distinguish
between real and fake images.
The goal of this discriminator is to help improve the quality of synthesized results
such that the trained discriminator is fooled by unrealistic images.

We first propose a local $\mathcal D$ for the missing region which determines whether the synthesized contents in the missing region are real or not.
Compared with Figure~\ref{fig:step}(c), the network with local $\mathcal D$ (shown in Figure~\ref{fig:step}(d)) begins to help generate details of missing contents with sharper boundaries.
It encourages the generated object parts to be semantically valid.
However, its limitations are also obvious due to the locality.
First, the local loss can neither regularize the global structure of a face, nor guarantee the statistical consistency within and outside the masked regions.
Second, while the generated new pixels are conditioned on their surrounding contexts, a local $\mathcal D$ can hardly generate a direct impact outside the masked regions during the back propagation, due to the unpooling structure of the decoder.
Consequently, the inconsistency of pixel values along region boundaries is obvious.

Therefore, we introduce another global $\mathcal D$ to determine the faithfulness of an entire image. The fundamental idea is that the newly generated contents should not only be realistic,
but also consistent to the surrounding contexts.
From Figure~\ref{fig:step}(e), the network with additional global $\mathcal D$
greatly alleviates the inconsistent issue and further enforce the generated contents
to be more realistic.
We note that the architecture of two discriminators are similar to~\cite{radford-2015-dcGAN}.

\subsection{Semantic Regularization}
\begin{figure}[t]
	\centering
	\small
	{
		\begin{tabular}{c@{\hspace{0.01\linewidth}}c@{\hspace{0.01\linewidth}}c@{\hspace{0.01\linewidth}}c@{\hspace{0.01\linewidth}}c@{\hspace{0.01\linewidth}}c@{\hspace{0.01\linewidth}}c@{\hspace{0.01\linewidth}}c@{\hspace{0.01\linewidth}}c}

			\includegraphics[width = .235\linewidth]{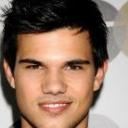} &
			\includegraphics[width = .235\linewidth]{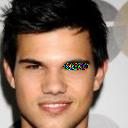} &
			\includegraphics[width = .235\linewidth]{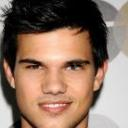} &
			\includegraphics[width = .235\linewidth]{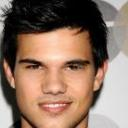} & \\
			
			\includegraphics[width = .235\linewidth]{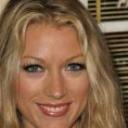} &
			\includegraphics[width = .235\linewidth]{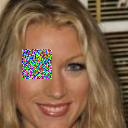} &
			\includegraphics[width = .235\linewidth]{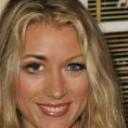} &
			\includegraphics[width = .235\linewidth]{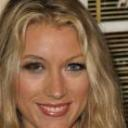} & \\
			
			{(a) original }& {(b) masked input}& {(c) w/o parsing}& {(d) w/ parsing} \\
			
		\end{tabular}
	}
	\caption{Comparison between the result of models without and with the parsing regularization.}
	
	\label{fig:parsing}
\end{figure}

With a generator and two discriminators, our model can be regarded as a variation of the original GAN~\cite{goodfellow-2014-GAN} model that is conditioned on contexts (e.g., non-mask regions).
%
However as a bottleneck, the GAN model tends to generate independent facial components that are likely not suitable to the original subjects with respect to facial expressions and parts shapes, as shown in Figure~\ref{fig:step}(e).
%
%
The top one is with big weird eyes and the bottom one
contains two asymmetric eyes.
Furthermore, we find the global $\mathcal D$ is not effective in ensuring
the consistency of fine details in the generated image.
For example, if only one eye is masked, the generated eye
does not fit well with another unmasked one.
We show another two examples in Figure~\ref{fig:parsing}(c) where the generated eye is obviously asymmetric to the unmasked one
although the generated eye itself is already realistic.
Both cases indicate that more regularization is needed to encourage the generated faces to have similar high-level distributions with the real faces.

Therefore we introduce a semantic parsing network to further enhance the harmony of the generated contents and existing pixels.
%
The parsing network is an autoencoder which bears some resemblance to
the semantic segmentation method~\cite{JMYang-2016-contour}.
The parsing result of the generated image is compared with the one of the original image.
As such, the generator is forced to learn where to generate
features with more natural shape and size.
In Figure~\ref{fig:step}(e)-(f) and Figure~\ref{fig:parsing}(c)-(d), we show the
generated images between models without and with the smenatic regularization.
%

\subsection{Objective Function}
\ignore{
The original GAN model~\cite{goodfellow-2014-GAN} is used to generate a new image from a noise vector.
It is supervised by a single adversarial loss that enforces the model to generate visually pleasing results.
The proposed model is different with~\cite{goodfellow-2014-GAN} in two aspects.
First, the known pixels outside the random mask need to be fitted well to the generated contents.
Second, the filled contents should contain consistent and correct object part information.
}

We first introduce a reconstruction loss  $L_r$ to the generator, which is the
$L_2$ distance between the network output and the original image.
%
With the $L_r$ only, the generated contents tend to be blurry and smooth as shown in Figure~\ref{fig:step}(c).
The reason is that since the $L_{2}$ loss penalizes outliers heavily, and
the network is encouraged to smooth across various hypotheses to avoid large penalties.


By using two discriminators, we employ the adversarial loss which is a reflection of how the generator can maximally fool the discriminator and how well the discriminator can distinguish between real and fake.
It is defined as
\begin{dmath}\label{formula2}
L_{a_i} = \min \limits_{\mathcal G}\max \limits_{\mathcal D} ~\mathcal{E}_{x \sim p_{data}(x)}[\log \mathcal D(x)]+\mathcal{E}_{z\sim p_z(z)}[\log (1-\mathcal D(\mathcal G(z)))],
\end{dmath}
where $p_{data}(x)$ and $p_z(z)$ represent the distributions of noise variables $z$ and real data $x$.
The two discriminative networks $\{a_1,a_2\}$ share the same definition of the loss function.
The only difference is that the local discriminator only provides training signals (loss gradients) for the missing region while the global discriminator back-propagates loss gradients across the entire image.

In the parsing network, the loss  $L_{p}$
is the simple pixel-wise softmax loss~\cite{FCN-2015-FCN,JMYang-2016-contour}.
The overall loss function is defined by
\begin{equation}\label{formula3}
L = L_{r} + \lambda_{1}L_{a_1} + \lambda_{2}L_{a_2} + \lambda_{3}L_{p},
\end{equation}
where $\lambda_{l}$, $\lambda_{2}$ and $\lambda_{3}$ are the weights
to balance the effects of different losses.

\subsection{Training Neural Networks}

To effectively train our network, we use the curriculum strategy~\cite{bengio-2009-curriculum} by gradually increasing the difficulty level and network scale.
The training process is scheduled in three stages.
First, we train the network using the reconstruction loss to obtain blurry contents.
Second, we fine-tune the network with the local adversarial loss.
The global adversarial loss and semantic regularization are incorporated at the last stage, as shown in Figure~\ref{fig:step}.
Each stage prepares features for the next one to improve, and hence greatly increases
the effectiveness and efficiency of network training.
For example, in Figure~\ref{fig:step}, the reconstruction stage (c)
restores the rough shape of the missing eye although the contents are blurry.
Then local adversarial stage (d) then generates more details to make the eye region visually realistic, and the global adversarial stage (e) refines the whole image to ensure that the appearance is consist around the boundary of the mask.
The semantic regularization (f) finally further enforces more consistency between components and let the generated result to be closer to the actual face.
When training with the adversarial loss, we use a method similar to~\cite{radford-2015-dcGAN} especially to avoid the case when the discriminator is too strong at the beginning of the training process.

\begin{figure}[t]
\centering
\small
{
\begin{tabular}{c@{\hspace{0.01\linewidth}}c@{\hspace{0.01\linewidth}}c@{\hspace{0.01\linewidth}}c@{\hspace{0.01\linewidth}}c@{\hspace{0.01\linewidth}}c@{\hspace{0.01\linewidth}}c@{\hspace{0.01\linewidth}}c@{\hspace{0.01\linewidth}}c}
\includegraphics[width = .145\linewidth]{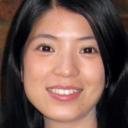} &
\includegraphics[width = .145\linewidth]{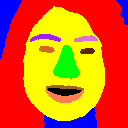} &

\hspace{1pt}\vrule\hspace{0.8pt}

\includegraphics[width = .145\linewidth]{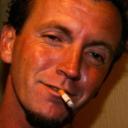} &
\includegraphics[width = .145\linewidth]{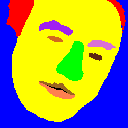} &

\hspace{1pt}\vrule\hspace{0.8pt}

\includegraphics[width = .145\linewidth]{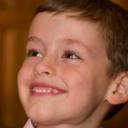} &
\includegraphics[width = .145\linewidth]{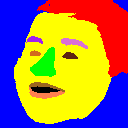} & \\

\includegraphics[width = .145\linewidth]{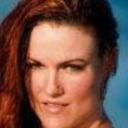} &
\includegraphics[width = .145\linewidth]{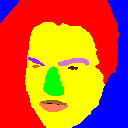} &

\hspace{1pt}\vrule\hspace{0.8pt}

\includegraphics[width = .145\linewidth]{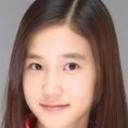} &
\includegraphics[width = .145\linewidth]{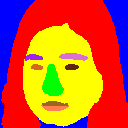} &

\hspace{1pt}\vrule\hspace{0.8pt}

\includegraphics[width = .145\linewidth]{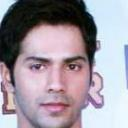} &
\includegraphics[width = .145\linewidth]{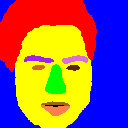} & \\

\end{tabular}
}
\caption{~Examples of our parsing results on Helen test dataset (top) and CelebA test dataset (bottom). In each panel, all pixels in the face image (left) are classified as one of 11 labels which are shown in different colors (right).}
\label{fig:parsing2}
\end{figure}

\section{Experimental Results}

\begin{figure*}[t]
\centering
\footnotesize
{
\begin{tabular}{c@{\hspace{0.01\linewidth}}c@{\hspace{0.01\linewidth}}c@{\hspace{0.01\linewidth}}c@{\hspace{0.01\linewidth}}c@{\hspace{0.01\linewidth}}c@{\hspace{0.01\linewidth}}c@{\hspace{0.01\linewidth}}c@{\hspace{0.01\linewidth}}c@{\hspace{0.01\linewidth}}c}

\includegraphics[width = .098\linewidth]{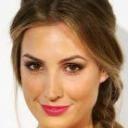} &
\includegraphics[width = .098\linewidth]{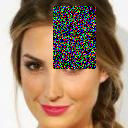} &
\includegraphics[width = .098\linewidth]{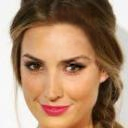} &

\hspace{1pt}\vrule\hspace{0.8pt}

\includegraphics[width = .098\linewidth]{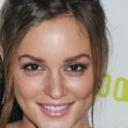} &
\includegraphics[width = .098\linewidth]{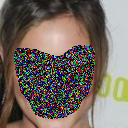} &
\includegraphics[width = .098\linewidth]{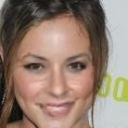} &

\hspace{1pt}\vrule\hspace{0.8pt}

\includegraphics[width = .098\linewidth]{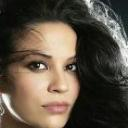} &
\includegraphics[width = .098\linewidth]{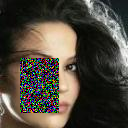} &
\includegraphics[width = .098\linewidth]{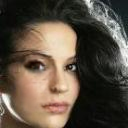} & \\

\includegraphics[width = .098\linewidth]{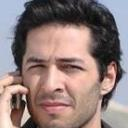} &
\includegraphics[width = .098\linewidth]{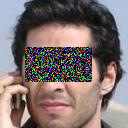} &
\includegraphics[width = .098\linewidth]{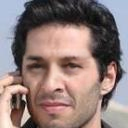} &

\hspace{1pt}\vrule\hspace{0.8pt}

\includegraphics[width = .098\linewidth]{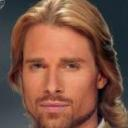} &
\includegraphics[width = .098\linewidth]{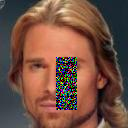} &
\includegraphics[width = .098\linewidth]{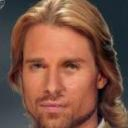} &

\hspace{1pt}\vrule\hspace{0.8pt}

\includegraphics[width = .098\linewidth]{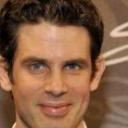} &
\includegraphics[width = .098\linewidth]{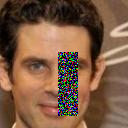} &
\includegraphics[width = .098\linewidth]{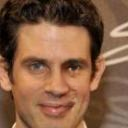} & \\

\includegraphics[width = .098\linewidth]{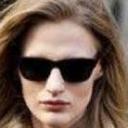} &
\includegraphics[width = .098\linewidth]{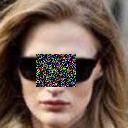} &
\includegraphics[width = .098\linewidth]{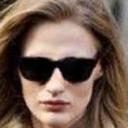} &

\hspace{1pt}\vrule\hspace{0.8pt}

\includegraphics[width = .098\linewidth]{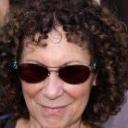} &
\includegraphics[width = .098\linewidth]{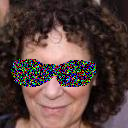} &
\includegraphics[width = .098\linewidth]{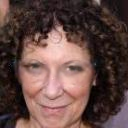} &

\hspace{1pt}\vrule\hspace{0.8pt}

\includegraphics[width = .098\linewidth]{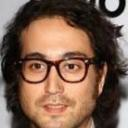} &
\includegraphics[width = .098\linewidth]{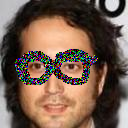} &
\includegraphics[width = .098\linewidth]{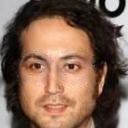} & \\

\includegraphics[width = .098\linewidth]{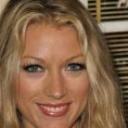} &
\includegraphics[width = .098\linewidth]{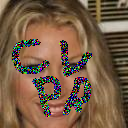} &
\includegraphics[width = .098\linewidth]{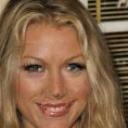} &

\hspace{1pt}\vrule\hspace{0.8pt}

\includegraphics[width = .098\linewidth]{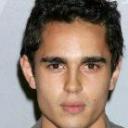} &
\includegraphics[width = .098\linewidth]{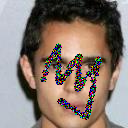} &
\includegraphics[width = .098\linewidth]{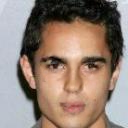} &

\hspace{1pt}\vrule\hspace{0.8pt}

\includegraphics[width = .098\linewidth]{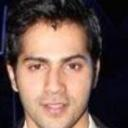} &
\includegraphics[width = .098\linewidth]{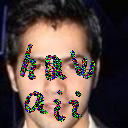} &
\includegraphics[width = .098\linewidth]{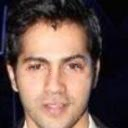} & \\

\end{tabular}
}
\caption{Face completion results on the CelebA~\cite{liu-2015-celebA} test dataset. In each panel from left to right: original images, masked inputs, our completion results.}
\label{fig:face_results_128}
\end{figure*}

We carry out
extensive experiments to demonstrate the ability of our model to synthesize the
missing contents on face images.
The hyper-parameters (e.g., learning rate) for the network training are set
as suggested in~\cite{wang-2016-SSGAN}.
To balance the effects of different losses, we use $\lambda_{l}=300$, $\lambda_{2}=300$ and $\lambda_{3}=0.005$ in our experiments.
%
%

\subsection{Datasets}

We use the CelebA~\cite{liu-2015-celebA} dataset to learn and evaluate our model. It consists of 202,599 face images and each face image is cropped, roughly aligned by the position of two eyes, and rescaled to $128\times128\times3$ pixels.
We follow the standard split with 162,770 images for training, 19,867 for validation and 19,962 for testing.
We set the mask size as $64\times64$ for training to guarantee that at least one essential facial component is missing.
%
If the mask only covers smooth regions with a small mask size, it will not drive the model to learn semantic representations.
To avoid over-fitting, we do data augmentation that includes flipping, shift, rotation (+/- 15 degrees) and scaling.
During the training process, the size of the mask is fixed but the position is randomly selected.
As such, the model is forced to learn the whole object in an holistic manner
instead of a certain part only.
%

\subsection{Face Parsing}
Since face images in the CelebA~\cite{liu-2015-celebA} dataset do not have segment labels, we use the Helen face dataset~\cite{ParsingHelen-2012-interactive} to train a face parsing network for regularization.
The Helen dataset consists of 2,330 images and each face has 11 segment labels covering every main component of the face (e.g., hair, eyebrows, eyes) labelled by~\cite{ParsingHelenLabel-2013-exemplar}.
%
%
We roughly crop the face in each image with the size of 128$\times$128 first and then feed
it into the parsing network to predict the label for each pixel.
Our parsing network bears some resemblance to the semantic segmentation method~\cite{JMYang-2016-contour} and we mainly modify its last layer with 11 outputs.
We use the standard training/testing split and obtain a parsing model, which achieves the f-score of 0.851 with overall facial components on the Helen test dataset, compared to the state-of-the-art multi-objective based model~\cite{Liu_2015_CVPR}, with the corresponding f-score of 0.854.
This model can be further improved with more careful hyperparameter tuning but is currently sufficient to improve the quality of face completion.
%
Several parsing results on the Helen test images
are presented in Figure~\ref{fig:parsing2}.

%

Once the parsing network is trained, it remains fixed in our generation framework.
We first use the network on the CelebA training set to obtain the parsing results of originally unmasked faces as the ground truth, and compare them with the parsing on generated faces during training.
The parsing loss is eventually back-propagated to the generator to regularize face completion.
We show some parsing results on the CelebA dataset in Figure~\ref{fig:parsing2}.
The proposed semantic regularization can be regarded as measuring
the distance in feature space where the sensitivity to local image statistics can be achieved~\cite{Doso-2016-perceptual}.

\subsection{Face Completion}

\paragraph{Qualitative results.}
Figure~\ref{fig:face_results_128} shows
our face completion results on the CelebA test dataset.
In each test image, the mask covers at least one key facial components.
The third column of each panel shows our completion results are visually realistic and pleasing.
Note that during the testing, the mask does not need to be restricted as a $64\times64$ square mask, but the number of total masked pixels is suggested to be no more than $64\times64$ pixels.
We show typical examples with one big mask covering at least two face components (e.g., eyes, mouths, eyebrows, hair, noses) in the first two rows.
We specifically present more results on eye regions since they can better reflect how realistic of the newly generated faces are, with the proposed algorithm.
Overall, the algorithm can successfully complete the images with faces in side views, or partially/completely corrupted by the masks with different shapes and sizes.

We present a few examples in the third row where the real occlusion (e.g., wearing glasses) occurs.
%
As sometimes whether a region in the image is
occluded or not is subjective, we give this option for users to assign
the occluded regions through drawing masks.
%
The results clearly show that our model is able to restore the partially masked eyeglasses, or remove the whole eyeglasses or just the frames by filling in realistic eyes and eyebrows.

\begin{figure}[t]
\centering
\footnotesize
{
\begin{tabular}{c@{\hspace{0.01\linewidth}}c@{\hspace{0.01\linewidth}}c@{\hspace{0.01\linewidth}}c@{\hspace{0.01\linewidth}}c@{\hspace{0.01\linewidth}}c@{\hspace{0.01\linewidth}}c@{\hspace{0.01\linewidth}}c@{\hspace{0.01\linewidth}}c@{\hspace{0.01\linewidth}}c}

\includegraphics[width = .23\linewidth]{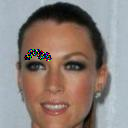} &
\includegraphics[width = .23\linewidth]{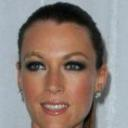} &

\hspace{0.8pt}\vrule\hspace{0.8pt}

\includegraphics[width = .23\linewidth]{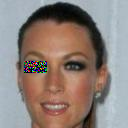} &
\includegraphics[width = .23\linewidth]{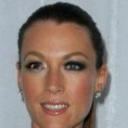} & \\

\includegraphics[width = .23\linewidth]{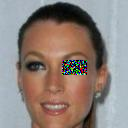} &
\includegraphics[width = .23\linewidth]{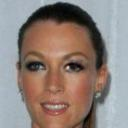} &

\hspace{0.8pt}\vrule\hspace{0.8pt}

\includegraphics[width = .23\linewidth]{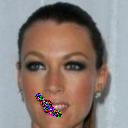} &
\includegraphics[width = .23\linewidth]{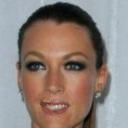} & \\

\includegraphics[width = .23\linewidth]{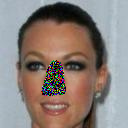} &
\includegraphics[width = .23\linewidth]{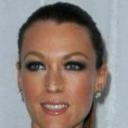} &

\hspace{0.8pt}\vrule\hspace{0.8pt}

\includegraphics[width = .23\linewidth]{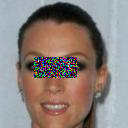} &
\includegraphics[width = .23\linewidth]{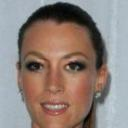} & \\

\end{tabular}
}
\caption{Face part completion. In each panel, left: masked input, right: our completion result.}
\label{fig:part}
\end{figure}

In the last row, we present examples with multiple, randomly drawn masks, which are closer to real-world application scenarios.
Figure~\ref{fig:part} presents completion results where different key parts (e.g., eyes, nose, and mouth) of the same input face image are masked.
It shows that our completion results are consistent and realistic
regardless of the mask shapes and locations.

\begin{figure}[t]
	\centering
	\footnotesize
	{
		\begin{tabular}{c@{\hspace{0.01\linewidth}}c@{\hspace{0.01\linewidth}}c@{\hspace{0.01\linewidth}}c@{\hspace{0.01\linewidth}}c@{\hspace{0.01\linewidth}}c@{\hspace{0.01\linewidth}}c@{\hspace{0.01\linewidth}}c@{\hspace{0.01\linewidth}}c@{\hspace{0.01\linewidth}}c}

			\includegraphics[width = .151\linewidth]{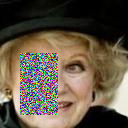} &
			\includegraphics[width = .151\linewidth]{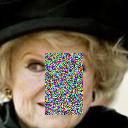} &
			\includegraphics[width = .151\linewidth]{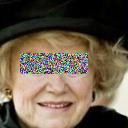} &
			\includegraphics[width = .151\linewidth]{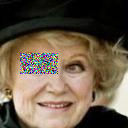} &
			\includegraphics[width = .151\linewidth]{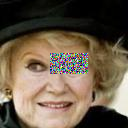} &
			\includegraphics[width = .151\linewidth]{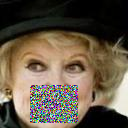} & \\
			
			{(a) O1}& {(b) O2}& {(c) O3}& {(d) O4}& {(e) O5}& {(f) O6}\\
			
		\end{tabular}
	}
	\caption{~Simulate face occlusions happened in real scenario with different masks O1-O6. From left to right: left half, right half, two eyes, left eye, right eye, and lower half. }
	\label{fig:occlusion}
\end{figure}

\paragraph{Quantitative results.}
In addition to the visual results, we also perform quantitative evaluation using three metrics on the CelebA test dataset (19,962 images).
The first one is the peak signal-to-noise ratio (PSNR) which directly measures the difference in pixel values.
The second one is the structural similarity index (SSIM) that estimates the holistic similarity between two images.
Lastly we use the identity distance measured by the OpenFace
toolbox~\cite{amos-2016-openface}
to determine the high-level semantic similarity of two faces.
These three metrics are computed between the completion results obtained by different methods and the original face images. The results are shown in Table~\ref{table:quantitative1}-\ref{table:quantitative3}.
Specifically, the stepwise contribution of each component is shown from the 2nd to the 5th column of each table, where M1-M5 correspond to five different settings of our own model in Figure~\ref{fig:step} and O1-O6 are six different masks for evaluation as shown in Figure~\ref{fig:occlusion}.
%

%
We then compare our model with the ContextEncoder~\cite{Deepak-2016-ContextEncoder} (CE).
%
%
Since the CE model is originally not trained for faces, we retrain the CE model on the CelebA dataset for fair comparisons.
As the evaluated masks O1-O6 are not in the image center, we use the \emph{inpaintRandom} version of their code and mask 25\% pixels masked in each image. Finally we also replace the non-mask region of the output with original pixels.
The comparison between our model (M4) and CE in 5th and 6th column show that our model performs generally better than the CE model, especially on large masks (e.g., O1-O3, O6).
In the last column, we show that the poisson blending~\cite{PB-2003-poisson} can further improve the performance.

Note that we obtain relatively higher PSNR and SSIM values when using the reconstruction loss (M1) only but it does not imply better qualitative results, as shown in Figure~\ref{fig:step}(c).
These two metrics simply favor smooth and blurry results.
We note that the model M1 performs poorly as it hardly recovers anything and is unlikely to preserve the identity well, as shown in Table~\ref{table:quantitative3}.

\begin{figure}[t]
\begin{center}
\includegraphics[height = .3\linewidth]{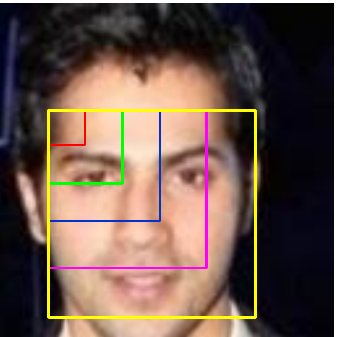}
\includegraphics[height = .5\linewidth]{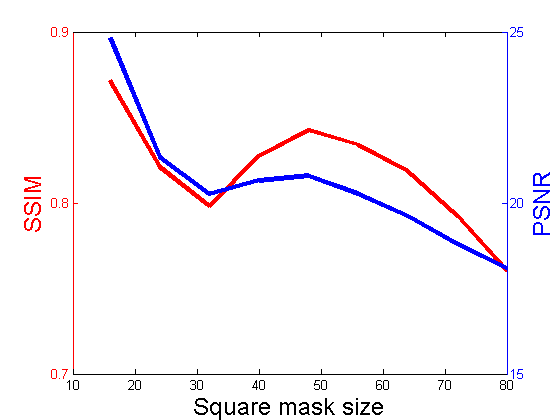}
\end{center}
\caption{~Evaluations on different square mask sizes of our final completion model (M5). The curve shows the average performance over all face images in the CelebA test dataset.}
\label{fig:masksize}
\end{figure}

Although the mask size is fixed as $64\times 64$ during the training, we test different sizes, ranging from 16 to 80 with a step of 8, to evaluate the generalization ability of our model.
Figure~\ref{fig:masksize} shows quantitative results.
The performance of the proposed model gradually drops with the increasing mask size,
which is expected as the larger mask size indicates more uncertainties in pixel values.
But generally our model performs well for smaller mask sizes (smaller than 64).
%
We observe a local minimum around the medium size (e.g., 32).
It is because that the medium size mask is mostly likely to occlude only part of the component (e.g., half eye). It is found in experiments that generating a part of the component is
more difficult than synthesizing new pixels for the whole component.
Qualitative results of different size of masking are presented in Figure~\ref{fig:face_results_128}.

\begin{table}[t]
\center
\small
\caption{~Quantitative evaluations in terms of SSIM at six different masks O1-O6. Higher values are better. }

\begin{tabular}{|c|c|c|c|c|c|c|}
\hline
~ & ~M1~ & ~M2~ & ~M3~ & ~M4~ &~CE~ & ~M5~ \\
\hline
O1 & 0.798 & 0.753 & 0.782 & 0.804 & 0.772 & \textbf{0.824}  \\
\hline
O2 & 0.805  & 0.763 & 0.787 & 0.808 & 0.774 & \textbf{0.826}\\
\hline
O3 & 0.723  & 0.675 & 0.708 & 0.731 & 0.719 & \textbf{0.759}\\
\hline
O4 & 0.747  & 0.701 & 0.741 & 0.759 & 0.754 & \textbf{0.789}\\
\hline
O5 & 0.751  & 0.706 & 0.732 & 0.755 & 0.757 & \textbf{0.784}\\
\hline
O6 & 0.807 & 0.764 & 0.808 & 0.824 & 0.818 & \textbf{0.841}\\
\hline
\end{tabular}
\label{table:quantitative1}
\end{table}

\begin{table}[t]
\center
\small
\caption{~Quantitative evaluations in terms of PSNR at six different masks O1-O6. Higher values are better.}
\begin{tabular}{|c|c|c|c|c|c|c|}
\hline
~ & ~M1~ & ~M2~ & ~M3~ & ~M4~ &~CE~ & ~M5~ \\
\hline
O1 & 18.9 & 17.8 & 18.9 & 19.4 & 18.6 & \textbf{20.0} \\
\hline
O2 & 18.7 & 17.9 & 18.7 & 19.3 & 18.4 & \textbf{19.8} \\
\hline
O3 & 17.9  & 17.2 & 17.7 & 18.3 & 17.9 & \textbf{18.8} \\
\hline
O4 & 18.6  & 17.7 & 18.5 & 19.1 & 19.0 & \textbf{19.7} \\
\hline
O5 & 18.7  & 17.6 & 18.4 & 18.9 & 19.1 & \textbf{19.5} \\
\hline
O6 & 18.8  & 17.3 & 19.0 & 19.7 & 19.3 & \textbf{20.2} \\
\hline
\end{tabular}
\label{table:quantitative2}
\end{table}

\begin{table}[t]
\center
\small
\caption{~Quantitative evaluations in terms of identity distance at six different masks O1-O6. Lower values are better.}
\begin{tabular}{|c|c|c|c|c|c|c|}
\hline
~ & ~M1~ & ~M2~ & ~M3~ & ~M4~ &~CE~ & ~M5~ \\
\hline
O1 & 0.763 & 0.775 & 0.694 & 0.602 & 0.701 & \textbf{0.534}\\
\hline
O2 & 1.05  & 1.02 & 0.894 & 0.838 & 0.908 & \textbf{0.752}\\
\hline
O3 & 0.781  & 0.693 & 0.674 & 0.571 & 0.561 & \textbf{0.549} \\
\hline
O4 & 0.310  & 0.307 & 0.265 & 0.238 & 0.236 & \textbf{0.212}  \\
\hline
O5 & 0.344  & 0.321 & 0.297 & 0.256 & 0.251 & \textbf{0.231} \\
\hline
O6 & 0.732  & 0.714 & 0.593 & 0.576 & 0.585 & \textbf{0.541}\\
\hline
\end{tabular}
\label{table:quantitative3}
\end{table}

\begin{figure}[t]
\centering
\small
{
\begin{tabular}{c@{\hspace{0.01\linewidth}}c@{\hspace{0.01\linewidth}}c@{\hspace{0.01\linewidth}}c@{\hspace{0.01\linewidth}}c@{\hspace{0.01\linewidth}}c@{\hspace{0.01\linewidth}}c@{\hspace{0.01\linewidth}}c@{\hspace{0.01\linewidth}}c@{\hspace{0.01\linewidth}}c@{\hspace{0.01\linewidth}}c@{\hspace{0.01\linewidth}}c}

\includegraphics[width = .264\linewidth]{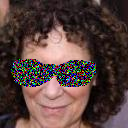} &
\includegraphics[width = .264\linewidth]{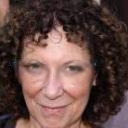} &
\includegraphics[width = .264\linewidth]{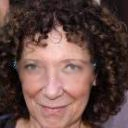} & \\

\includegraphics[width = .264\linewidth]{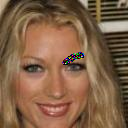} &
\includegraphics[width = .264\linewidth]{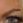} &
\includegraphics[width = .264\linewidth]{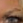} & \\

\end{tabular}
}
\caption{Completion results under different noisy inputs. The generated contents are all semantically plausible but with different appearances. Check the shape of the eye (top) and the right side of the eyebrow (bottom). Moreover, the difference is also reflected by shades and tints.  Note that as constrained by the contexts, the variations on appearance is unlikely to be too diverse.}
\label{fig:latent}
\end{figure}

\paragraph{Traversing in latent space.}
The missing region, although semantically constrained by the remaining pixels in an image,
accommodates different plausible appearances as shown in Figure~\ref{fig:latent}.
%
%
We observe that when the mask is filled with different noise, all the generated contents are
semantically realistic and consistent, but their appearances varies.
%
%
This is different from the context encoder~\cite{Deepak-2016-ContextEncoder},
where the mask is filled with zero values
and thus the model only renders single completion result.

It should be noted that under different input noise, the variations of our generated contents are unlikely to be as large as those in the original GAN~\cite{goodfellow-2014-GAN,radford-2015-dcGAN} model which is able to generate completely different faces. This is mainly due to the constraints from the contexts (i.e., non-mask regions).
%
For example, in the second row of Figure~\ref{fig:latent} with only one eyebrow masked, the generated eyebrow is restricted to have the similar shape and size and reasonable position with the other eyebrow.
Therefore the variations on the appearance of the generated eyebrow are mainly reflected at some details, such as the shade of the eyebrow.

\begin{figure*}[t]
\centering
\footnotesize
{
\begin{tabular}{c@{\hspace{0.01\linewidth}}c@{\hspace{0.01\linewidth}}c@{\hspace{0.01\linewidth}}c@{\hspace{0.01\linewidth}}c@{\hspace{0.01\linewidth}}c@{\hspace{0.01\linewidth}}c@{\hspace{0.01\linewidth}}c@{\hspace{0.01\linewidth}}c@{\hspace{0.01\linewidth}}c}

\includegraphics[height = .24\linewidth]{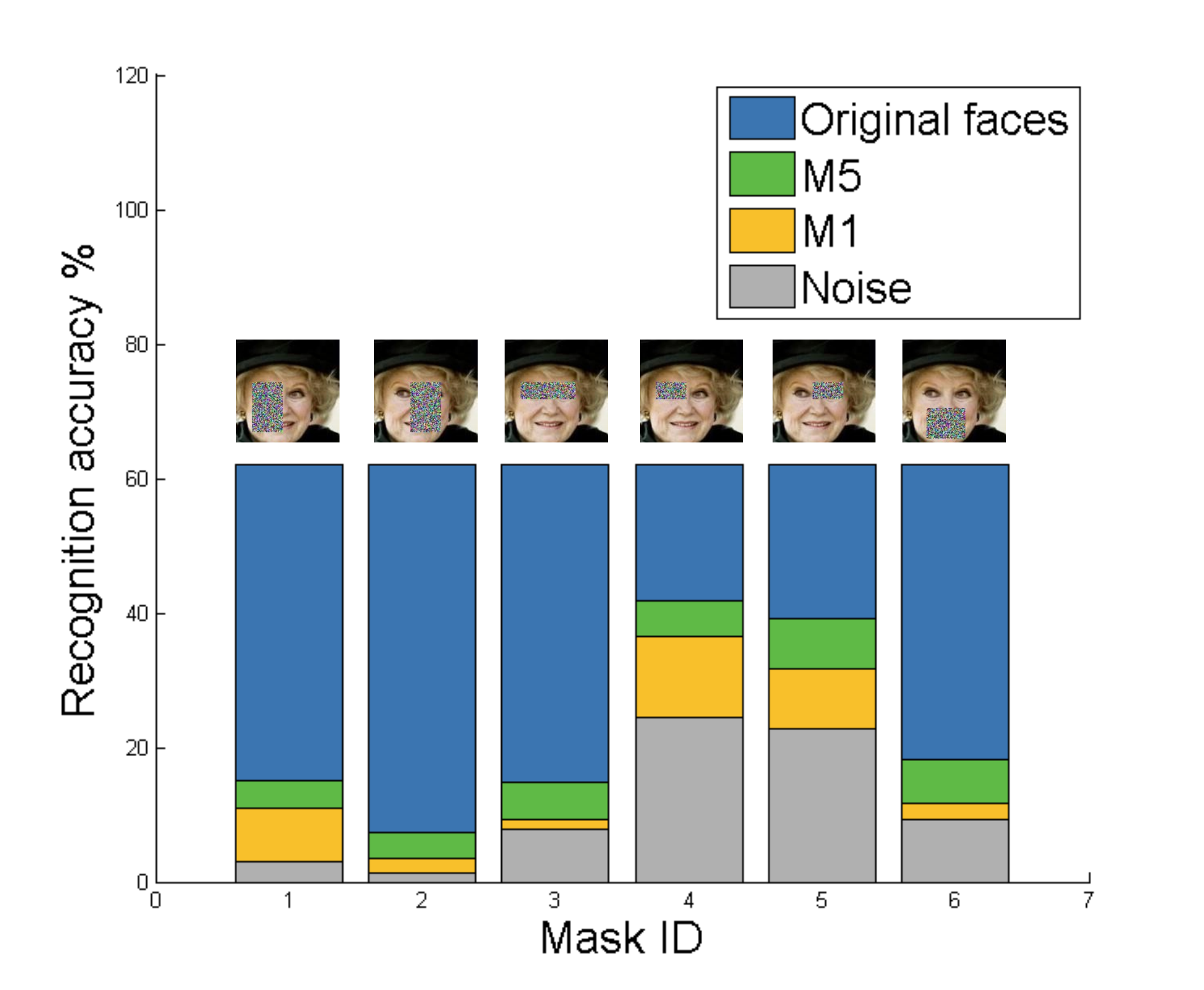} &
\includegraphics[height = .24\linewidth]{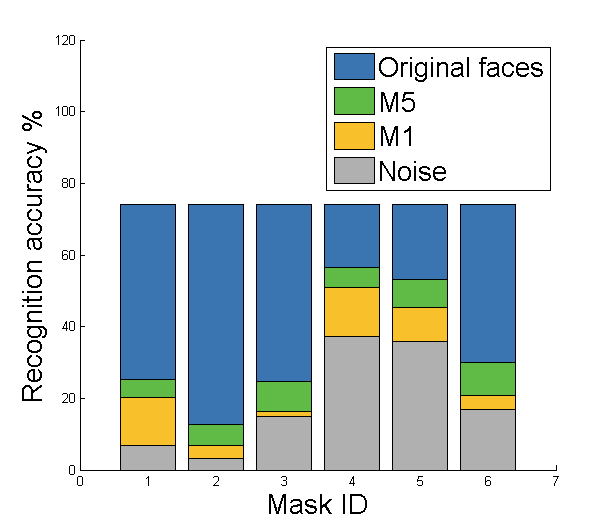} &
\includegraphics[height = .24\linewidth]{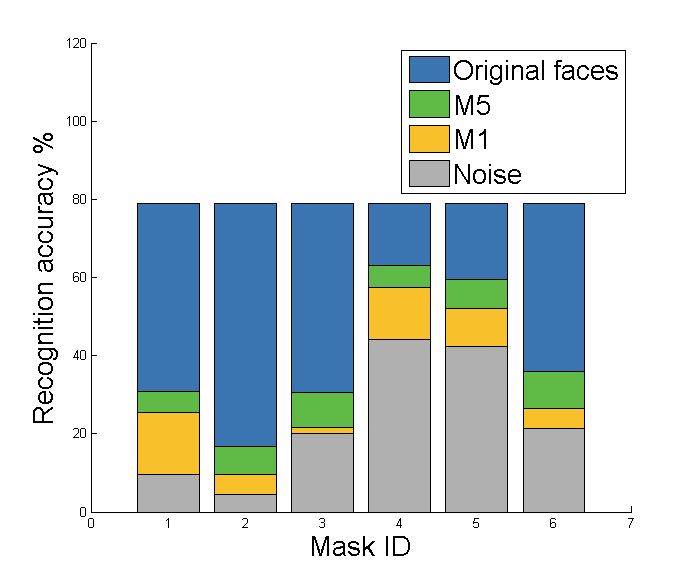} & \\

{ (a) Top1} & { (b) Top3}& { (c) Top5}\\

\end{tabular}
}
\caption{~Recognition accuracy comparisons on masked (or occluded) faces. Given a masked probe face, we first complete it and then use it to search examples of the same identity in the gallery. We report the Top1, Top3, and Top5 recognition accuracy of three different completion methods. The accuracy by using the original unmasked probe face (blue) is treated as the standard to compare.}
\label{fig:recognition}
\end{figure*}

\subsection{Face recognition}

The identity distance in Table~\ref{table:quantitative3} partly reveals the network ability of preserving the identity information.
%
%
In order to test to what extent the face identity can be preserved across its different examples, we evaluate our completion results in the task of face recognition.
Note that this task simulates occluded face recognition, which is still an open problem in computer vision.
Given a probe face example, the goal of recognition is to find an example from the gallery set that belongs to the same identity.
We randomly split the CelebA~\cite{liu-2015-celebA} test dataset into the \emph{gallery} and \emph{probe} set, to make sure that each identity has roughly the same amount of images in each set.
Finally, we obtain the gallery and probe set with roughly 10,000 images respectively, covering about 1,000 identities.

We apply six masking types (O1-O6) for each probe image, as shown in Figure~\ref{fig:occlusion}.
The probe images are new faces restored by the generator.
These six masking types, to some extent, simulate the occlusions that possibly occurs
in real scenarios.
For example, masking two eyes mainly refers to the occlusion by glasses and masking lower half face matches the case of wearing the scarf.
Each completed probe image is matched against those in the gallery, and top ranked matches can be analyzed to measure recognition performance.
We use the OpenFace~\cite{amos-2016-openface} toolbox to find top $K$ nearest matches based on the identity distance and report the average top $K$ recognition accuracy over all probe images in Figure~\ref{fig:recognition}.

We carry out experiments with four variations of the probe image: the original one,
the completed one by simply filling random noise, by our reconstruction based model M1 and by our final model M5.
The recognition performance using original probe faces is regarded as the upper bound.
Figure~\ref{fig:recognition} shows that using the completed probe by our model M5 (green) achieves the closest performance to the upper bound (blue).
Although there is still a large gap between the performance of our M5 based recognition and the upper bound, especially when the mask is large (e.g., O1, O2),
the proposed algorithm makes significant improvement with the completion results
compared with that by either noise filling or the reconstruction loss ($L_{r}$).
We consider the identity-preserving completion to be an interesting direction to pursue.

\subsection{Limitations}

Although our model is able to generate semantically plausible and visually pleasing contents, it has some limitations.
The faces in the CelebA dataset are roughly cropped and aligned~\cite{liu-2015-celebA}.
We implement various data augmentation to improve the robustness of learning, but find our model still cannot handle some unaligned faces well.
We show one failure case in the first row of Figure~\ref{fig:limitation}.
The unpleasant synthesized contents indicate that the network does not recognize the position/orientation of the face and its corresponding components.
This issue can be alleviated with 3D data augmentation.

In addition, our model does not fully exploit the spatial correlations between adjacent pixels
as shown in the second row of Figure~\ref{fig:limitation}.
The proposed model fails to recover the correct color of the lip, which is originally painted with red lipsticks.
%
%
%
In our future work, we plan to investigate the usage of pixel-level recurrent neural network (PixelRNN~\cite{PixelRNN-ICML2016}) to address this issue.



\begin{figure}[t]
\centering
\footnotesize
{
\begin{tabular}{c@{\hspace{0.01\linewidth}}c@{\hspace{0.01\linewidth}}c@{\hspace{0.01\linewidth}}c@{\hspace{0.01\linewidth}}c@{\hspace{0.01\linewidth}}c@{\hspace{0.01\linewidth}}c@{\hspace{0.01\linewidth}}c@{\hspace{0.01\linewidth}}c@{\hspace{0.01\linewidth}}c}

\includegraphics[width = .264\linewidth]{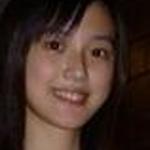} &
\includegraphics[width = .264\linewidth]{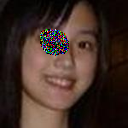} &
\includegraphics[width = .264\linewidth]{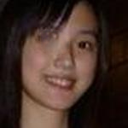} \\
\includegraphics[width = .264\linewidth]{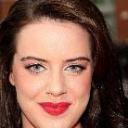} &
\includegraphics[width = .264\linewidth]{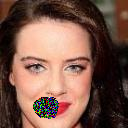} &
\includegraphics[width = .264\linewidth]{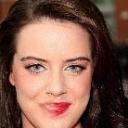} & \\

\end{tabular}
}
\caption{Model limitations. Top: our model fails to generate the eye for an unaligned face. Bottom: it is still hard to generate the semantic part with right attributes (e.g., red lipsticks).}
\label{fig:limitation}
\end{figure}
\ignore{
\begin{figure}[t]
\centering
\footnotesize
{
\begin{tabular}{c@{\hspace{0.01\linewidth}}c@{\hspace{0.01\linewidth}}c@{\hspace{0.01\linewidth}}c@{\hspace{0.01\linewidth}}c@{\hspace{0.01\linewidth}}c@{\hspace{0.01\linewidth}}c@{\hspace{0.01\linewidth}}c@{\hspace{0.01\linewidth}}c@{\hspace{0.01\linewidth}}c}

\includegraphics[width = .15\linewidth]{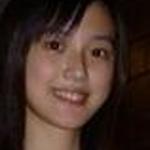} &
\includegraphics[width = .15\linewidth]{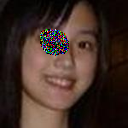} &
\includegraphics[width = .15\linewidth]{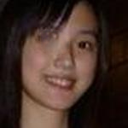} &

\hspace{1pt}\vrule\hspace{0.8pt}

\includegraphics[width = .15\linewidth]{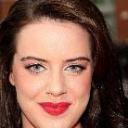} &
\includegraphics[width = .15\linewidth]{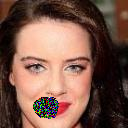} &
\includegraphics[width = .15\linewidth]{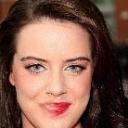} & \\

\end{tabular}
}
\caption{Model limitations. Left: our model fails to generate the eye for an unaligned face. Right: it is still hard to generate the semantic part with right attributes (e.g., painted red lipsticks).}
\label{fig:limitation}
\end{figure}
}

\section{Conclusion}

In this work we propose a deep generative network for face completion.
%
The network is based on a GAN, with an autoencoder as the generator, two adversarial loss functions (local and global) and a semantic regularization as the discriminators.
The proposed model can successfully synthesize semantically valid and visually plausible contents for the missing facial key parts from random noise.
%
%
Both qualitative and quantitative experiments show that our model generates the completion results of high perceptual quality and is quite flexible to handle a variety of maskings or occlusions (e.g., different positions, sizes, shapes).

\paragraph{\bf Acknowledgment.} This work is supported in part by the NSF CAREER Grant \#1149783, gifts from Adobe and Nvidia.

{\small
\bibliographystyle{ieee}
\bibliography{egbib}
}

\end{document}